\documentclass[10pt,twocolumn,letterpaper]{article}

\usepackage[pagenumbers]{wacv} 

\usepackage[accsupp]{axessibility}

\usepackage{graphicx}
\usepackage{epsfig}
\usepackage{amsmath}
\usepackage{amssymb}
\usepackage{booktabs}
\usepackage{graphicx}
\usepackage{multirow}
\usepackage{multicol}

\usepackage{tabularx}
\usepackage{adjustbox}
\usepackage{array}
\usepackage{multirow}
\usepackage{url}
\usepackage{booktabs}
\usepackage{array}
\usepackage{lipsum}
\usepackage{makecell} 
\usepackage{xcolor}
\usepackage{caption, subcaption}

\usepackage[pagebackref,breaklinks,colorlinks]{hyperref}

\usepackage[capitalize]{cleveref}
\crefname{section}{Sec.}{Secs.}
\Crefname{section}{Section}{Sections}
\Crefname{table}{Table}{Tables}
\crefname{table}{Tab.}{Tabs.}

\def\wacvPaperID{132} 
\def\confName{WACV}
\def\confYear{2024}

\def\ours{CAILA\,}
\def\openset{{\large{$\circ$}}}
\def\closedset{{\large{$\bullet$}}}
\usepackage{pifont}
\newcommand{\cmark}{\ding{51}}%
\def\eg{\emph{e.g.} }

\def\etal{\emph{et al.} }
\newcommand{\ttilde}{$\sim$}

\renewcommand\vec{\mathbf}
\DeclareMathOperator*{\argmax}{arg\,max}

\usepackage{enumitem}
\definecolor{lightgreen}{HTML}{E2F0D9}

\newcommand\blfootnote[1]{%
\begingroup 
\renewcommand\thefootnote{}\footnote{#1}%
\addtocounter{footnote}{-1}%
\endgroup 
}

\begin{document}

\title{CAILA: Concept-Aware Intra-Layer Adapters \\ for Compositional Zero-Shot Learning}

\author{Zhaoheng Zheng \quad Haidong Zhu \quad Ram Nevatia\\
Viterbi School of Engineering\\
University of Southern California\\
{\tt\small \{zhaoheng.zheng, haidongz, nevatia\}@usc.edu}
}
\maketitle

\begin{abstract}

In this paper, we study the problem of Compositional Zero-Shot Learning (CZSL), which is to recognize novel attribute-object combinations with pre-existing concepts. Recent researchers focus on applying large-scale Vision-Language Pre-trained (VLP) models like CLIP with strong generalization ability. However, these methods treat the pre-trained model as a black box and focus on pre- and post-CLIP operations, which do not inherently mine the semantic concept between the layers inside CLIP. We propose to dive deep into the architecture and insert adapters, a parameter-efficient technique proven to be effective among large language models, into each CLIP encoder layer. We further equip adapters with concept awareness so that concept-specific features of ``object'', ``attribute'', and ``composition'' can be extracted. We assess our method on four popular CZSL datasets, MIT-States, C-GQA, UT-Zappos, and VAW-CZSL, which shows state-of-the-art performance compared to existing methods on all of them.
  
\end{abstract}

\section{Introduction}
\blfootnote{Code will be available at  \href{https://github.com/zhaohengz/CAILA}{github.com/zhaohengz/CAILA}}

When facing a novel concept such as a \textit{large castle}, humans can deconstruct individual components (\textit{large} and \textit{castle}) from familiar concepts (\textit{large bear, old castle}) to comprehend the new composition. Such task of recognizing new attribute-object compositions based on a set of observed pairs is Compositional Zero-Shot Learning (CZSL) \cite{misra2017red}, a sine qua non for an intelligent entity. However, the inherent challenge in CZSL lies in the capacity to identify unobserved novel compositions without compromising the recognition of previously observed combinations. Conventional approaches \cite{li2020symmetry,atzmon2020causal,li2022siamese,mancini2022learning,saini2022disentangling,purushwalkam2019task,wei2019adversarial,yang2020learning,nan2019recognizing,nagarajan2018attributes,misra2017red,wang2019tafe,naeem2021learning} often suffer from training biases. Even though recent methods employ large-scale Vision-Language Pre-training (VLP) models with strong generalization ability, e.g., CLIP \cite{radford2021learning}, to accommodate this issue, they simply treat VLP models as frozen black box encoders and fail to exploit the potential of VLP models. Thus, here, we explore how to more effectively extract and utilize the knowledge embedded in pre-trained vision-language models for the recognition of novel attribute-object compositions.

\begin{figure}[t]
     \centering
     \hfill
     \begin{subfigure}[b]{0.32\linewidth}
         \centering
         \includegraphics[width=\textwidth]{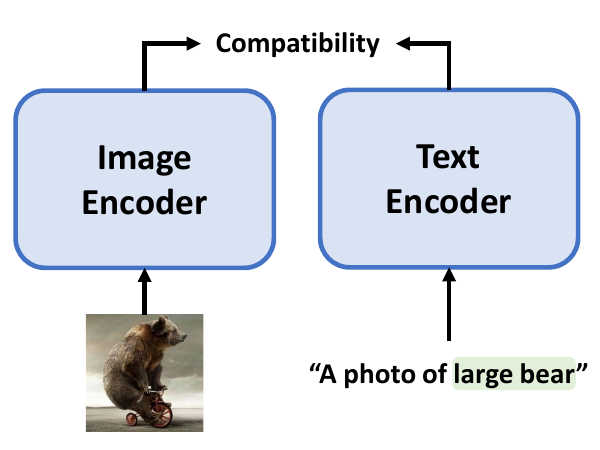}
         \caption{Prompt Tuning \cite{nayak2023learning}}
         \label{fig:prompt}
     \end{subfigure}
     \begin{subfigure}[b]{0.32\linewidth}
         \centering
         \includegraphics[width=\textwidth]{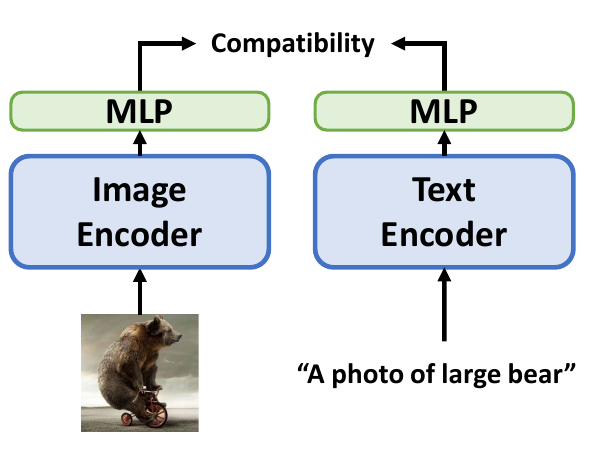}
         \caption{CLIP-Adapter \cite{gao2021clip}}
         \label{fig:clip-adapter}
     \end{subfigure}
     \begin{subfigure}[b]{0.32\linewidth}
         \centering
         \includegraphics[width=\textwidth]{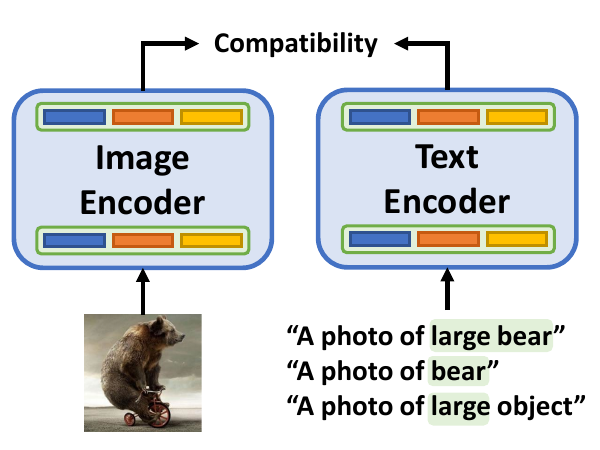}
         \caption{\ours (Ours)}
         \label{fig:ours-thumb}
     \end{subfigure}
     \hfill
     \\
    \begin{subfigure}[b]{0.5\linewidth}
         \centering
         \includegraphics[width=\textwidth]{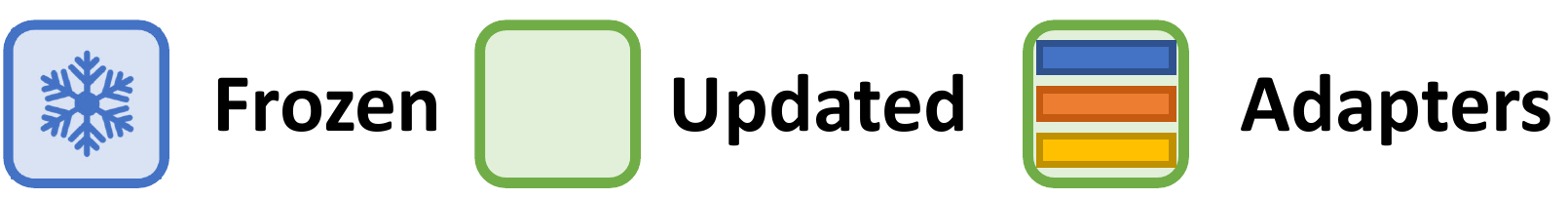}
    \end{subfigure}
    \\
    \begin{subfigure}[b]{\linewidth}
         \centering
    \includegraphics[width=\textwidth]{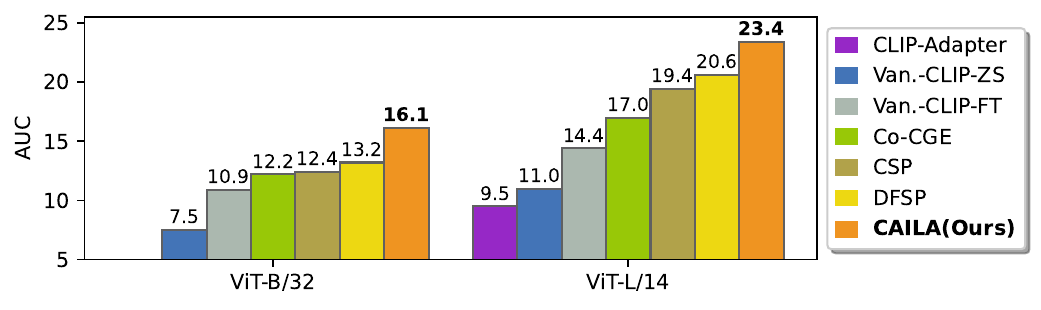}
    \caption{Comparisons of AUC on MIT-States \cite{isola2015discovering} with different backbones}
    \label{fig:auc-thumb}
    \end{subfigure}
        \caption{Illustrations of \ours and previous CLIP-based baselines.
        \ours has adapters integrated into both CLIP encoders and thus better transfers the knowledge from CLIP to CZSL, resulting in significant performance boosts compared with other CLIP-based baseline methods. ``Van.-CLIP'' refers to models with vanilla CLIP architecture. \colorbox{lightgreen}{\texttt{Prompts}} highlighted in green are set to be learnable parameters.}
        \label{fig:teaser}
    \vspace{-15pt}
\end{figure}

More specifically, to adapt VLP models for CZSL, some researchers apply prompt-tuning \cite{nayak2023learning,zhou2022conditional,zhou2022learning} or fine-tune the model with extra adaptation layers \cite{gao2021clip} on the top of CLIP. However, prompt-tuning methods, depicted in Figure~\ref{fig:teaser}(a), only learn trainable prompts, while CLIP-Adapter, shown in Figure~\ref{fig:teaser}(b), only adds external modules outside CLIP. Both strategies abstain from altering the fundamental CLIP encoder, consequently retaining CLIP as a static black box. Nayak \textit{et al.} \cite{nayak2023learning} have shown that exhaustively fine-tuning CLIP falls short of attaining practicable performance. Thus, we argue that properly optimizing features across layers through a task-specific design is critical to effectively harnessing the knowledge embedded in CLIP. A feasible CLIP-based CZSL should: i) have task-specific designs for CZSL; ii) be capable of extracting concept-specific features related to compositions and individual primitives.  

Hence, we propose \textbf{\ours}, \textbf{C}oncept-\textbf{A}ware \textbf{I}ntra-\textbf{L}ayer \textbf{A}dapters, that satisfy the given prerequisites and substantiate its superiority, as shown in Fig.~\ref{fig:teaser}(d), compared with other CLIP-based methods. Fig.~\ref{fig:teaser}(c) highlights the difference between \ours and other VLP-based methods. Instead of prompt tuning or fully fine-tuning, we adopt adapters \cite{houlsby2019parameter} to transfer knowledge from VLP models while avoiding strong training biases. 

Moreover, given that adapters are low-overhead components, it is feasible to employ a variety of adapters to extract concept-wise representations. More specifically, \ours integrates a group of adapters into each layer of both encoders; each group possesses concept-specific components to extract knowledge corresponding to particular concepts, including attributes, objects, and compositions. To merge features extracted by various concept-aware adapters, we propose the \textbf{M}ixture-of-\textbf{A}dapters (MoA) mechanism for both vision and yrcy encoder. In addition, the property that CAILA can extract concept-specific features allows us to further propose Primitive Concept Shift, which generates additional vision embeddings by combining the attribute feature from one image and the object feature from another for a more comprehensive understanding.

We evaluate our approach on three popular CZSL datasets: MIT-States \cite{isola2015discovering}, C-GQA \cite{naeem2021learning} and UT-Zappos \cite{finegrained2014,semjitter2017}, under both closed world and open world settings. We also report the performance of \ours in closed world on VAW-CZSL\cite{saini2022disentangling}, a newly released benchmark. Our experiments show that, in both scenarios, our model beats the state-of-the-arts over all benchmarks following the generalized evaluation protocol \cite{purushwalkam2019task}, by significant margins.

To summarize, our contributions are as follows: (i) We propose CAILA, which is the first model exploring CZSL-oriented designs with CLIP models to balance model capacity and training bias robustness; (ii) we design the Mixture-of-Adapter (MoA) mechanism to fuse the knowledge from concept-aware adapters and improve the generalizability; (iii) we further enrich the training data and exploit the power of CAILA through Primitive Concept Shifts; (iv) we conduct extensive experiments in exploring the optimal setup for \ours on CZSL. Quantitative experiments show that our model outperforms the  SOTA by significant margins in both closed world and open world, on all benchmarks.

\section{Related Works}

\textbf{Zero-Shot Learning (ZSL).} Unlike conventional fully-supervised learning, ZSL requires models to learn from side information without observing any visual training samples \cite{lampert2013attribute}. The side information comes from multiple non-visual resources such as attributes \cite{lampert2013attribute}, word embeddings \cite{wang2018zero,socher2013zero}
, and text descriptions \cite{reed2016learning}. Notably, Zhang \etal \cite{zhang2017learning} propose to learn a deep embedding model bridging the seen and the unseen, while \cite{chen2021free,xian2018feature,zhu2018generative} investigate generative models that produce features for novel categories. Moreover, \cite{wang2018zero,kampffmeyer2019rethinking} integrate Graph Convolution Networks (GCN) \cite{KipfW17} to better generalize over unseen categories. 

\textbf{Compostional Zero-Shot Learning (CZSL).} Previous CZSL approaches are built with pre-trained image encoders, \eg ResNet and separate word embeddings, \eg GloVe \cite{pennington2014glove}. More specifically, Li \etal \cite{li2020symmetry} investigate the symmetrical property between objects and attributes, while Atzmon \etal \cite{atzmon2020causal} study the casual influence between the two. Moreover, Li \etal \cite{li2022siamese} construct a Siamese network with contrastive learning to learn better object/attribute prototypes. On the other hand, joint representations of compositions can be leveraged in multiple ways. \cite{purushwalkam2019task} utilizes joint embeddings to control gating functions for the modular network, while \cite{wei2019adversarial,yang2020learning,nan2019recognizing,nagarajan2018attributes} treat them as categorical centers in the joint latent space.
Furthermore, some approaches \cite{misra2017red,wang2019tafe,naeem2021learning,ruis2021independent,mancini2022learning} directly take compositional embeddings as classifier weights, while OADis \cite{saini2022disentangling} disentangles attributes and objects in the visual space. 

\textbf{Parameter-Efficient Tuning. } Recent research on large scale pre-training models \cite{radford2021learning,jia2021scaling,hu2022scaling,li2021align, goenka2022fashionvlp} has achieved superior performance on various downstream tasks, compared with regular approaches. Various works \cite{houlsby2019parameter,sung2022vl,karimi2021compacter} show that tuning adapters \cite{houlsby2019parameter} on the language side yields comparable results with fully fine-tuned variants, while Chen \etal \cite{chen2022vision} investigate the adaptation of image encoders on dense prediction tasks. For CZSL, a few models \cite{zhou2022learning, nayak2023learning} leverage the knowledge of CLIP through prompt tuning \cite{lester2021power} , while Gao \etal \cite{gao2021clip} attach a post-processor to CLIP for knowledge transfer. Though these methods show strong performance on CZSL against regular models, they treat the CLIP model as a black box and keep it completely frozen. In \ours, we open up the CLIP black box by integrating intra-layer adapters to both image and text encoders.

\section{Approach}

\label{sec:approach}

The problem of CZSL can be formulated as follows. We denote the training set by $\mathcal{T} = \{(x, y) | x \in \mathcal{X}, y \in \mathcal{Y}_{s}\}$, where $\mathcal{X}$ contains images represented in the RGB color space and $\mathcal{Y}_{s}$ is a set of seen composition labels which are available during the training phase. Each label $y = (a, o)$ is a pair of attribute $a \in \mathcal{A}$ and object category $o \in \mathcal{O}$. When testing, CZSL expects models to predict a set of unseen compositions $\mathcal{Y}_{u}$ that is mutually exclusive with training labels $\mathcal{Y}_{s}$: $\mathcal{Y}_{s} \cup \mathcal{Y}_{u} = \varnothing$. Note that $\mathcal{Y}_{s}$ and $\mathcal{Y}_{u}$ share the same set of $\mathcal{A}, \mathcal{O}$, while CZSL assumes that each $a \in \mathcal{A}$ or $o \in \mathcal{O}$ exists in the training set and only the composition $(a, o) \in \mathcal{Y}_{u}$ is novel. Following \cite{purushwalkam2019task,xian2018zero,naeem2021learning}, we focus on generalized CZSL, where the test set contains both seen and unseen labels, formally denoted by $\mathcal{Y}_{test} = \mathcal{Y}_{s} \cup \mathcal{Y}_{u}$.

Most recent works \cite{naeem2021learning,purushwalkam2019task,atzmon2020causal} study the generalized CZSL problem under the \textit{closed world} setting, where $\mathcal{Y}_{test}$ is a subset of the complete composition set $\mathcal{Y}: \mathcal{A} \times \mathcal{O}$. The \textit{closed world} setting assumes that $\mathcal{Y}_{u}$ are known during testing and thus greatly reduce the size of the search space. On the contrary, Mancini \etal \cite{mancini2021open} argue that such constraint should not be applied to the search space and introduce the \textit{open world} setting, where models are required to search over the complete set of compositions, formally $\mathcal{Y}_{s} \cup \mathcal{Y}_{u} = \mathcal{Y}$. In this paper, we investigate the problem in both \textit{closed world} and \textit{open world}.

\subsection{Compatibility Estimation Pipeline}
\label{sec:comp-est}
As different attributes can lead to significant appearance shifts even inside the same object category,  performing attribute and object predictions separately may be ineffective. Hence, we  model attribute-object compositions jointly and learn a combined estimation function to measure the compatibility of input image $x$ and query composition $(a, o)$. In addition, we let the model estimate attribute and object compatibilities as auxiliary sub-tasks during training. 

The estimation of composition compatibility is represented as $\mathcal{C}(x, a, o): \mathcal{X} \times \mathcal{A} \times \mathcal{O} \rightarrow \mathbb{R}$. It contains two components: The image feature extractor $\mathcal{F}_C: \mathbb{R}^{H\times W \times 3} \rightarrow \mathbb{R}^{d}$ and the text embedding generator $\mathcal{G}: \mathcal{A} \times \mathcal{O} \rightarrow \mathbb{R}^d$. Note that $d$ denotes the number of channels that each representation has. Given an image $x$ and a composition $(a, o)$, the compatibility score is defined as the dot product of $\mathcal{F}_C(x)$ and $\mathcal{G}(a, o)$, formally
\begin{equation}
    \mathcal{C}(x, a, o) = \mathcal{F}_C(x) \cdot \mathcal{G}(a, o).
    \label{eqn:est}
\end{equation}

Furthermore, as CZSL requires models to recognize novel pairs composed of known attributes and objects, it is important for a model to possess the capability of primitive feature extraction that is disentangled with training compositions. Thus, we make our model extract features corresponding to primitives and estimate the compatibility between vision features and text representations during training. Similar to Eqn.~\ref{eqn:est}, we have 
\begin{equation}
    \mathcal{C}(x, a) = \mathcal{F}_{A}(x) \cdot \mathcal{G}_{A}(a), \ \mathcal{C}(x, o) = \mathcal{F}_{O}(x) \cdot \mathcal{G}_{O}(o).
\end{equation}
All three compatibility scores contribute independently to the loss function, while $\mathcal{C}(x, a, o)$ is leveraged during inference.
More specifically, our framework learns separate representations through CAILA discussed in Sec.~\ref{sec:CAILA} and conducts knowledge fusion through \textbf{M}ixture-of-\textbf{A}dapters (MoA), which will be covered in Sec.~\ref{sec:moa}.

Following \cite{radford2021learning}, we create a prompt template similar to \texttt{"a photo of [CLASS]"} for each compatibility estimation sub-task. For composition compatibility, we feed the text encoder with \texttt{"a photo of [ATTRIBUTE] [OBJECT]"}; We use \texttt{"a photo of [ATTRIBUTE] object"} and \texttt{"a photo of [OBJECT]"} for attribute and object compatibilities, respectively. Similar to \cite{nayak2023learning}, we only make \texttt{[CLASS]} prompts trainable. For both encoders $\mathcal{F}$ and $\mathcal{G}$, we take the output hidden state of the \textbf{\texttt{[CLS]}} token as the representation.

\begin{figure*}[t]
    \centering
    \includegraphics[width=.95\textwidth]{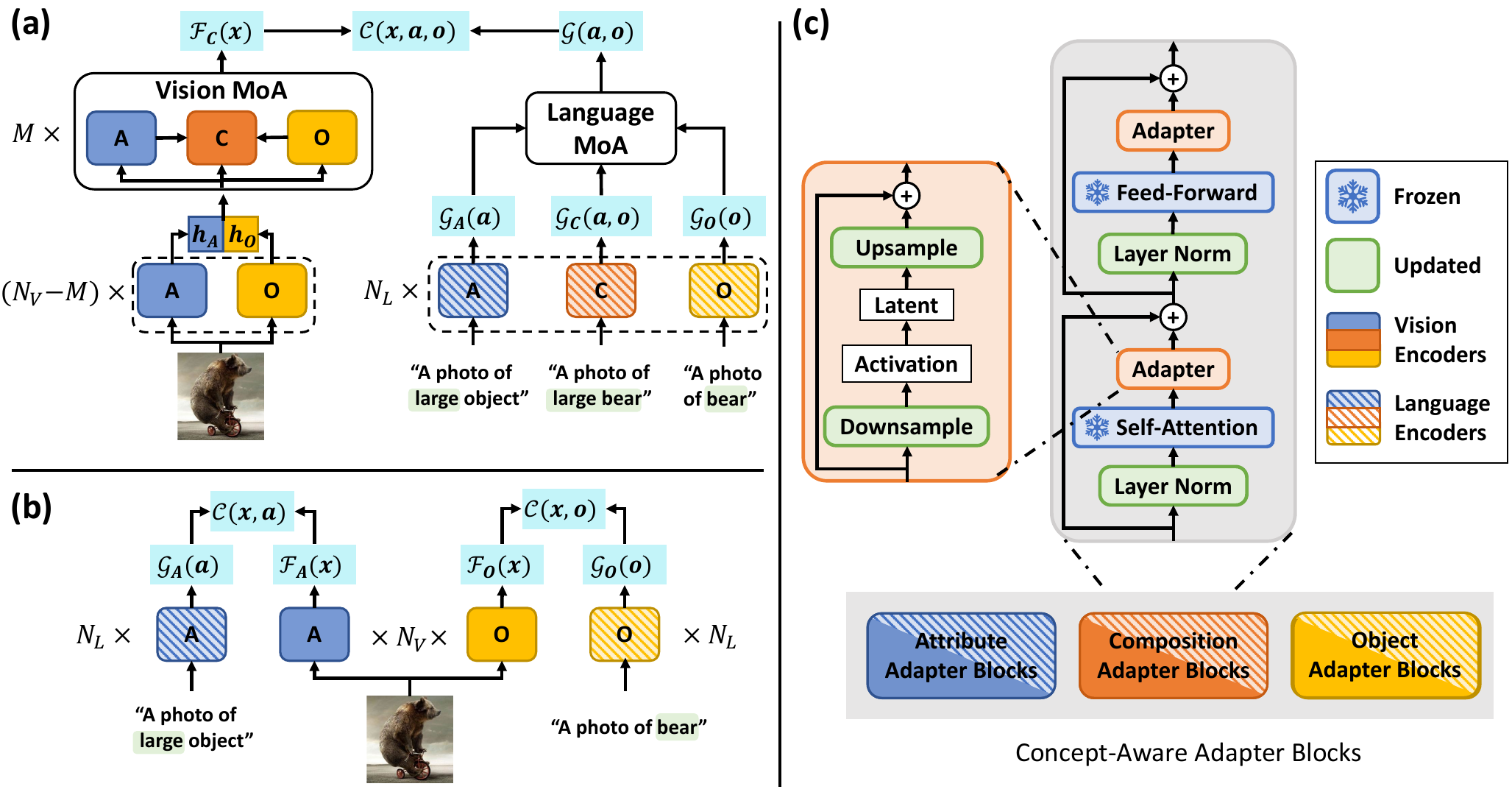}
    \caption{An overview of \ours: (a) The main composition compatibility estimation pipeline; (b) Auxiliary sub-tasks on primitive compatibility during training; (c) The structure of \ours  layers. Our model extracts concept-specific features by learning different adapters and fuses them through the Mixture-of-Adapters (MoA) mechanism. Note that for each layer of encoders in (a) and (b), the weights of encoding blocks of the same concept are shared. $N_V, N_L$ and $M$ indicate numbers of layers.}
    \label{fig:pipeline}
    \vspace{-10pt}
\end{figure*}

\subsection{Concept-Aware Intra-Layer Adapters}
\label{sec:CAILA}

Though CLIP-based CZSL approaches \cite{nayak2023learning,gao2021clip,zhou2022learning} have achieved significant improvements compared with earlier methods \cite{misra2017red,naeem2021learning,mancini2021open,purushwalkam2019task,nayak2023learning}, the CLIP encoder is considered as a black box and no modifications are made to improve its generalizability. Thus, we propose to improve CLIP-based CZSL models in both modalities with \ours, Concept-Aware Intra-Layer Adapters. 

As shown in Fig.~\ref{fig:pipeline} (a)(b), we take the CLIP image encoder as $\mathcal{F}$ and the text encoder as $\mathcal{G}$, while adding concept awareness to both encoders when estimating compatibilities of different concepts. Fig.~\ref{fig:pipeline} (c) demonstrates how adapters are integrated into a regular transformer encoding block. For each encoding block, we add adapters behind the frozen self-attention layer and the feed-forward network. More specifically, given the input hidden state $\vec{h}$ of an adapter, we compute the latent feature $\vec{z}$ by the downsampling operator $f_{Down}$, followed by the activation function $\sigma$. The output $\vec{h'}$ of an adapter is obtained by upscaling $\vec{z}$ and summing it with $\vec{h}$ through the skip connection. Formally, we have 
\begin{equation}
\vec{z} = \sigma(f_{Down}(\vec{h})), \quad \vec{h'} = f_{Up}(\vec{z}) + \vec{h},
\end{equation}
where both $f_{Down}$ and $f_{Up}$ are fully-connected layers.

To extract concept-specific features, at each depth level, we create three encoding blocks corresponding to attribute, object, and composition, respectively. As in Fig.~\ref{fig:pipeline}(c),  encoding blocks of at the same level share the same weights except for the adapter layers. Inputs from both modalities are processed by encoders equipped with different types of encoding blocks and features related to each of the three concepts are produced. During training, vision-language compatibility scores for ``attribute'', ``object'' and ``compositions'' are estimated. More specifically, encoders referred in Fig.~\ref{fig:pipeline}(a) and (b) are the same ones; There are not extra side encoders for auxiliary sub-tasks. 

\subsection{MoA: Mixture of Adapters}
\label{sec:moa}
\begin{figure}[t]
    \centering
    \includegraphics[width=\linewidth]{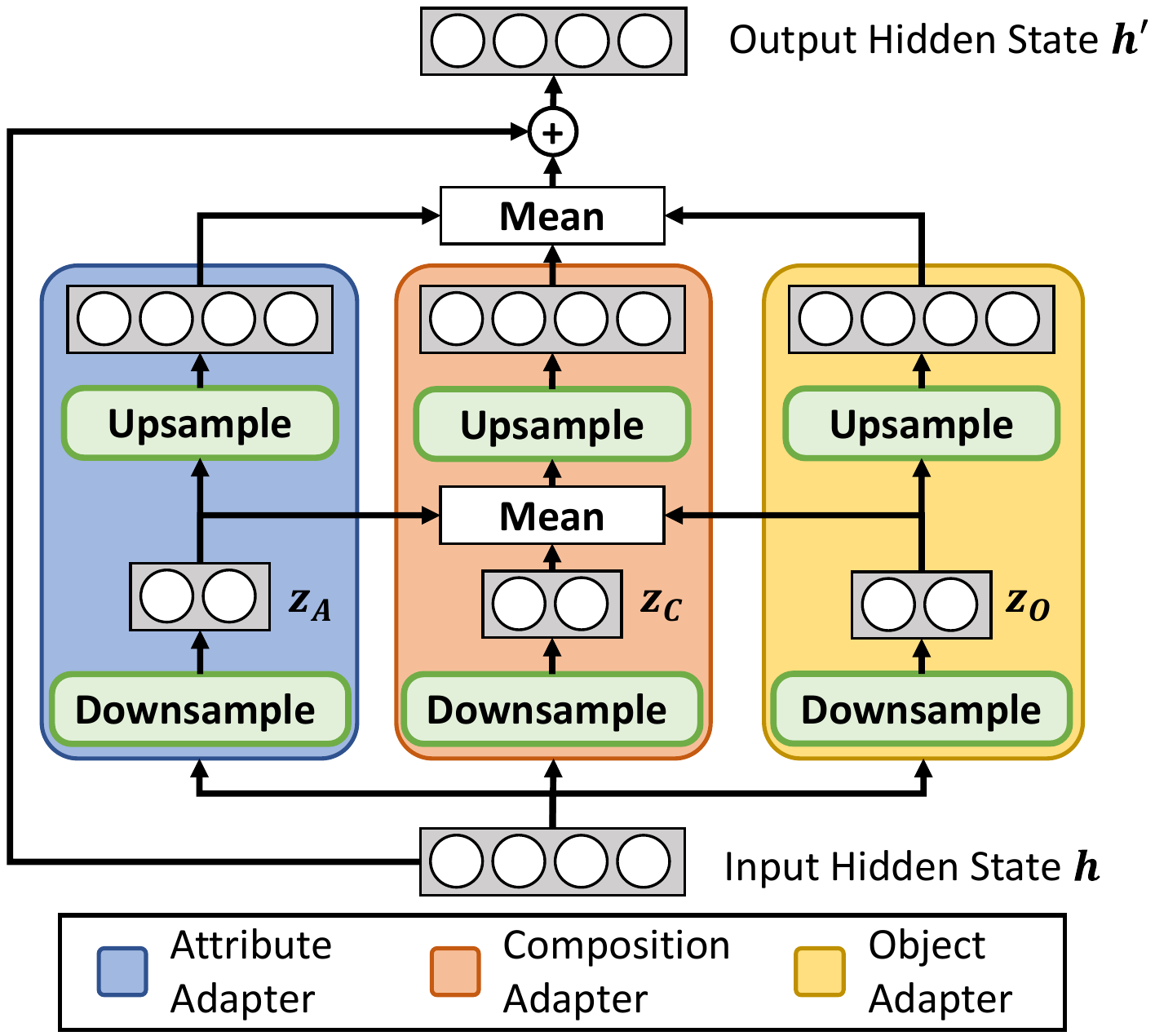}
    \caption{Details of our vision Mixture-of-Adapter. Latent features of each adapter, $\vec{z_A}, \vec{z_O}, \vec{z_C}$, are mixed, and further processed by the upsampling function to generate $\vec{h'_C}$. $\vec{h'_C}$ is joined with $\vec{h'_A}, \vec{h'_O}$ and input feature $\vec{h}$ for output.}
    \label{fig:vision-moa}
    \vspace{-18pt}
\end{figure}
To aggregate the knowledge extracted by adapters corresponding to attributes, objects, and compositions, we propose Mixture-of-Adapters mechanisms for both the vision side and language side of the encoder. 

On the vision side, we perform a two-stage feature extraction. As shown in Fig.~\ref{fig:pipeline} (a), for the first $N_V - M$ layers, we extract features related to the attribute ($\vec{h}_A$) and the object ($\vec{h}_O$)  through corresponding encoding blocks, which are further concatenated and processed by the trailing $M$ ternary MoA layers. An example of the vision MoA layer is shown in Fig.~\ref{fig:vision-moa}. Given the hidden state $h$, we extract latent features $\vec{z_A}$, $\vec{z_O}$ and $\vec{z_C}$ from the adapters. We then combine all three features and create $\vec{z'_C}$, followed by $f_{Up}$:
\begin{equation}
    \vec{z'_C} = \text{Avg}\big[\vec{z_A}, \vec{z_O}, \vec{z_C}\big], \quad \vec{h'_C} = f_{Up}(\vec{z'_C}).
    \label{eqn:mix-z}
\end{equation}
We further combine $\vec{h'_C}$ with outputs of attribute and object adapters, $\vec{h'_A}$ and $\vec{h'_O}$, to create the output:
\begin{equation}
    \vec{h'} = \text{Avg}\big[\vec{h'_A}, \vec{h'_O}, \vec{h'_C}\big] + \vec{h}.
    \label{eqn:mix-h}
\end{equation}
The output of the last mixture layer is L2-normalized and adopted as $\mathcal{F}_{C}(x)$ for compatibility estimation. Ablation study on this module is discussed in Sec.~\ref{sec:ablation}.

Unlike the vision side, where attributes and objects are deeply entangled within the same input image. On the language side, we can create disentangled language inputs through different prompt templates for attributes and objects separately. Thus, we adopt a simple mixture strategy for language adapters. We compute the compositional embedding through $N_L$ encoding blocks for the composition and combine it with primitive language embeddings:
\begin{equation}
    \mathcal{G}(a, o) = \text{Avg}\big[\mathcal{G}_A(a), \mathcal{G}_O(o), \mathcal{G}_C(a, o)\big].
\end{equation}

\subsection{Primitive Concept Shift on Image Embeddings}
\begin{figure}
    \centering
    \includegraphics[width=\linewidth]{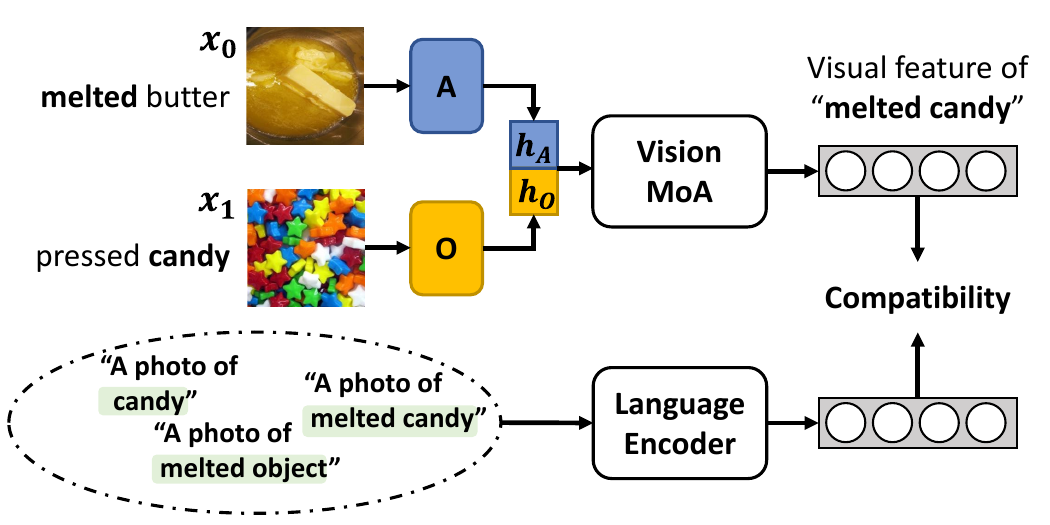}
    \caption{Illustrations of concept shift. We perform concept shift by combining the attribute (\textbf{melted}) feature from one image with the object (\textbf{candy}) feature to create an additional composition (\textbf{melted candy}) feature. Newly generated features are shuffled with regular samples during training.}
    \label{fig:shift}
    \vspace{-15pt}
\end{figure}
 Due to the limited diversity of training data, current CZSL models often suffer from training biases. As discussed in Sec.~\ref{sec:moa}, in addition to the composition-related feature, \ours extracts attribute- and object-oriented features during the first stage of $\mathcal{F}_{C}$. That motivates us to leverage these primitive-specific features to create additional embeddings for certain compositions. As it leads to changes in labels of original images, \eg from \texttt{melted butter} to \texttt{melted candy}, we call it \textit{primitive concept shift}.

Fig.~\ref{fig:shift} demonstrates the process of concept shift: Given one sample $x_0$ of \texttt{melted butter} and one sample $x_1$ of \texttt{pressed candy}, we create a new sample of \texttt{melted candy} in the feature space, by combining the attribute-oriented feature $\vec{h}_A$ of $x_0$ and the object-oriented feature $\vec{h}_O$ of $x_1$. The newly combined feature is further processed by vision MoA layers described in Sec.~\ref{sec:moa}, leading to an embedding representing \texttt{melted candy}. Such change can be viewed as an ``object shift'' from \texttt{melted butter} or an ``attribute shift'' from \texttt{pressed candy}. Thus, we name this process ``primitive concept shift''. In practice, we randomly pick a proportion of samples for shifting and ensure that the new label after shifting still lies in the training set. We discuss the effectiveness of the shifting in Sec~.\ref{sec:ablation}.

Although there are previous explorations \cite{li2022siamese,wei2019adversarial} in generating novel features in the latent space, our method is essentially novel from two aspects: i) Wei \etal \cite{wei2019adversarial} generate features directly from word embeddings, while our method leverages disentangled vision features that have richer and more diverse knowledge; ii) Li \etal \cite{li2022siamese} uses generated features to augment primitive vision encoders, while ours augments the entire model through \ours for both compositions and individual primitives.

\begin{table*}[t]
\centering

\resizebox{\textwidth}{!}{%
\begin{tabular}{c|r|rrrr|rrrr|rrrr}\toprule
& Closed World & \multicolumn{4}{c|}{\closedset{} MIT-States}          & \multicolumn{4}{c|}{\closedset{} C-GQA}                & \multicolumn{4}{c}{\closedset{} UT-Zappos }            \\
& Model & AUC ($\uparrow$) & HM ($\uparrow$)  & S ($\uparrow$)   & U ($\uparrow$) & AUC ($\uparrow$) & HM ($\uparrow$)  & S ($\uparrow$)   & U  ($\uparrow$)  & AUC ($\uparrow$) & HM ($\uparrow$)  & S ($\uparrow$)   & U ($\uparrow$) \\ 
\cmidrule(lr){1-14}

\multirow{6}{*}{\shortstack{Without \\ CLIP}}& CompCos \cite{mancini2021open}
& 4.5 & 16.4 & 25.3 & 24.6 & 2.6 & 12.4 & 28.1 & 11.2  & 28.7 & 43.1 & 59.8 & 62.5  \\
& ProtoProp \cite{ruis2021independent}  & - & - & - & - & 3.7 & 15.1 & 26.4 & 18.1 & 34.7 & 50.2 & 62.1 & 65.7  \\
& OADis \cite{saini2022disentangling} &  5.9 &  18.9 & 31.1 & 25.6 & -  & -  & -  & -  & 30.0 &  44.4 &  59.5 &  65.5\\
& SCEN \cite{li2022siamese} &  5.3 &  18.4 &  29.9 &  25.2 &  2.9 &  12.4 &  28.9 &  12.1 & 32.0 & 47.8 & 63.5 & 63.1 \\
& CGE \cite{naeem2021learning}
& 6.5 & 21.4 & 32.8 & 28.0 & 4.2 & 15.5 & 33.5 & 16.0  & 33.5 & \textbf{60.5} &64.5 & 71.5
\\
 & Co-CGE \cite{mancini2022learning} & {6.6} &  20.0 &  32.1 &  28.3 & 4.1 &  14.4 &  33.3 &  14.9 &  33.9 &  48.1 &  62.3 &  66.3 \\ 
  & CAPE \cite{khan2023learning} & 6.7 & 20.4 & 32.1 & 28.0 & 4.6 & 16.3 & 33.0 & 16.4 & 35.2 & 49.5 & 62.3 & 68.5 \\
 \cmidrule(lr){1-14}

\multirow{5}{*}{\shortstack{With \\ CLIP}} & CLIP-ZS \cite{radford2021learning} & 11.0 & 26.1 & 30.2 & 46.0 & 1.4 & 8.6 & 7.5 & 25.0 & 5.0 & 15.6 & 15.8 & 49.1 \\

& CoOp \cite{zhou2022learning} & 13.5 & 29.8 & 34.4 & 47.6 & 4.4 & 17.1 & 26.8 & 20.5 & 18.8 & 34.6 & 52.1 & 49.3 \\
& Co-CGE$^\dag$\cite{mancini2022learning} & 17.0 & 33.1 & 46.7 & 45.9 & 5.7 & 18.9 & 34.1 & 21.2 & 36.3 & 49.7 & 63.4 & 71.3 \\
& CSP \cite{nayak2023learning} & 19.4 & 36.3 & 46.6 & 49.9 & 6.2 & 20.5 & 28.8 & 26.8 & 33.0 & 46.6 & 64.2 & 66.2 \\
& DFSP \cite{lu2023decomposed} & 20.6 & 37.3 & 46.9 & 52.0 & 10.5 & 27.1 & 38.2 & 32.9 & 36.0 & 47.2 & 66.7 & 71.7 \\
\cmidrule(lr){2-14}
& \ours (Ours) &  \textbf{23.4} & \textbf{39.9} & \textbf{51.0} & \textbf{53.9} &
\textbf{14.8} & \textbf{32.7} & \textbf{43.9} & \textbf{38.5} &
\textbf{44.1} & 57.0 & \textbf{67.8} & \textbf{74.0} \\
\bottomrule

\end{tabular}%
}
\caption{Quantitative results on generalized CZSL in \textit{closed world}, all numbers are reported in percentage. S and U refer to best seen and unseen accuracy on the accuracy curve. CLIP-ZS refers to the vanilla CLIP model without fine-tuning. All CLIP-based models are run with ViT-L/14 and we conduct extensive experiments in Tab.~\ref{tab:clip}. \dag We run Co-CGE with similar CLIP features and report our best number of the model. Models published before CGE are omitted as their performances are inferior to current baselines.} 
\label{tab:closed}
\vspace{-10pt}
\end{table*}

\begin{table}[t]
\centering

\resizebox{\linewidth}{!}{%
\begin{tabular}{c|r|rrrr}\toprule
& Closed World & \multicolumn{4}{c}{\closedset{} VAW-CZSL}     \\
& Model & AUC ($\uparrow$) & HM ($\uparrow$)  & S ($\uparrow$)   & U ($\uparrow$)  \\ 
\cmidrule(lr){1-6}

\multirow{3}{*}{\shortstack{Without \\ CLIP}}& CompCos \cite{mancini2021open}& 5.6 & 14.2 & 23.9 & 18.0 \\
& OADis \cite{saini2022disentangling} &  6.1 & 15.2 & 24.9 & 18.7 \\
& CGE \cite{naeem2021learning} & 5.1 & 13.0 & 23.4 & 16.8\\
 \cmidrule(lr){1-6}

\multirow{5}{*}{\shortstack{With \\ CLIP}} & CLIP-ZS \cite{radford2021learning} & 2.6 & 11.9 & 12.8 & 27.8\\
& CSP \cite{nayak2023learning} & 8.5 & 23.3 & 31.9 & 33.6 \\
& DFSP \cite{lu2023decomposed} & 14.1 & 31.1 & 40.1 & 40.9\\
\cmidrule(lr){2-6}
& \ours (Ours) &  \bf 17.2 & \bf 34.6 & \bf 41.6 & \bf 49.2  \\
\bottomrule

\end{tabular}%
}
\caption{Quantitative results on generalized CZSL of VAW-CZSL in closed world, all numbers are reported in percentage.}

\label{tab:vaw-czsl}
\vspace{-10pt}
\end{table}

\subsection{Training and Testing}
\label{sec:train-test}

\textbf{Objective.} 
We optimize our model with a main loss on attribute-object compositions and auxiliary losses on attributes and objects. As our model only has access to seen compositions $Y_{s}$, we create our training objective upon $Y_{s}$ and ignore other compositions during training. More specifically, given an image $x$, we compute the compatibility score $\mathcal{C}(x, a, o), \mathcal{C}(x, a)$ and $\mathcal{C}(x, o)$ for all $(a, o) \in Y_{s}$. We then jointly optimize $\mathcal{F}$ and $\mathcal{G}$ by the cross-entropy loss with temperature:
\begin{align}
\begin{split}
\label{eqn:loss}
     & \mathcal{L} = \frac{-1}{|\mathcal{T}|}\sum_{i} 
    \bigg\{\log\frac{e^{[\mathcal{C}(x_i, a_i, o_i)/\tau_C]}}{\sum\limits_{j}e^{ [\mathcal{C}(x_i, a_j, o_j)/\tau_C]}} + \\ 
    & \log\frac{e^{[\mathcal{C}(x_i, a_i)/\tau_A]}}{\sum\limits_{j}e^{[\mathcal{C}(x_i, a_j)/\tau_A]}} + \log\frac{e^{[\mathcal{C}(x_i, o_i)/\tau_O]}}{\sum\limits_{j}e^{[\mathcal{C}(x_i, o_j)/\tau_O]}}\bigg\}.
\end{split}
\end{align}
Intuitively, the cross-entropy loss will force the model to produce a higher compatibility score when $(x, a, o)$ matches and lower the score when a non-label composition occurs. 

\textbf{Inference.} The generalized CZSL task requires models to perform recognition over a joint set of seen and unseen compositions. Thus, for each test sample $x$, we estimate the compatibility score between $x$ and every candidate $(a, o)$ inside the search space $\mathcal{Y}_{s} \cup \mathcal{Y}_{u}$. We predict the image $x$ as the composition that has the highest compatibility score:
\begin{equation}
\hat{y} = \argmax_{(a, o) \in \mathcal{Y}_s \cup \mathcal{Y}_u} \mathcal{C}(x, a, o)
\end{equation}
We apply the prediction protocol to all benchmarks.

\section{Experiments}

\subsection{Experiment Settings}
\label{sec:setting}

\begin{table*}[t]
\centering

\resizebox{\textwidth}{!}{%
\begin{tabular}{c|r|rrrr|rrrr|rrrr}\toprule
& Open World & \multicolumn{4}{c|}{\openset{} MIT-States}          & \multicolumn{4}{c|}{\openset{} C-GQA}                & \multicolumn{4}{c}{\openset{} UT-Zappos }            \\
& Model & AUC ($\uparrow$) & HM ($\uparrow$)  & S ($\uparrow$)   & U ($\uparrow$) & AUC ($\uparrow$) & HM ($\uparrow$)  & S ($\uparrow$)   & U  ($\uparrow$)  & AUC ($\uparrow$) & HM ($\uparrow$)  & S ($\uparrow$)   & U ($\uparrow$) \\ 
\cmidrule(lr){1-14}

\multirow{5}{*}{\shortstack{Without \\ CLIP}} & CompCos \cite{mancini2021open} & 0.8 & 5.8 & 21.4 & 7.0 & 0.43 & 3.3 & 26.7 & 2.2 & 18.5 & 34.5 & 53.3 & 44.6 \\
& CGE \cite{naeem2021learning} & 1.0 & 6.0 & 32.4 & 5.1 & 0.47 & 2.9 & 32.7 & 1.8 & 23.1 & 39.0 & 61.7 & 47.7 \\
& KG-SP \cite{karthik2022kg} & 1.3 & 7.4 & 28.4 & 7.5  & 0.78 & 4.7 & 31.5 & 2.9 & 26.5 & 42.3 & 61.8 & 52.1 \\
& Co-CGE$^{CW}$ \cite{mancini2022learning} & 1.1 & 6.4 & 31.1 & 5.8  & 0.53 & 3.4 & 32.1 & 2.0 & 23.1 & 40.3 & 62.0 & 44.3 \\
& Co-CGE$^{open}$ \cite{mancini2022learning} & 2.3 & 10.7 & 30.3 & 11.2 & 0.78 & 4.8 & 32.1 & 3.0 & 23.3 & 40.8 & 61.2 & 45.8 \\
\cmidrule(lr){1-14}
\multirow{6}{*}{\shortstack{With \\ CLIP}} & CLIP-ZS \cite{radford2021learning} & 3.0 & 12.8 & 30.1 & 14.3 & 0.27 & 4.0 & 7.5 & 4.6 & 2.2 & 11.2 & 15.7 & 20.6 \\
& CoOp (a)\cite{zhou2022learning} & 4.7 & 16.1 & 36.8 & 16.5 & 0.73 & 5.7 & 20.9 & 4.5 & 19.5 & 35.6 & 61.8 & 39.3 \\
& CoOp (b)\cite{zhou2022learning} & 2.8 & 12.3 & 34.6 & 9.3 & 0.70 & 5.5 & 21.0 & 4.6 & 13.2 & 28.9 & 52.1 & 31.5 \\
& Co-CGE$^\dag$\cite{mancini2022learning} & 5.6 & 17.7 & 38.1 & 20.0 & 0.91 & 5.3 & 33.2 & 3.9 & 28.4 & 45.3 & 59.9 & 56.2 \\
& CSP \cite{nayak2023learning} & 5.7 & 17.4 & 46.3 & 15.7 & 1.20 & 6.9 & 28.7 & 5.2 & 22.7 & 38.9 & 64.1 & 44.1 \\ 
& DFSP \cite{lu2023decomposed} & 6.8 & 19.3 & 47.5 & 18.5 & 2.40 & 10.4 & 38.3 & 7.2 & 30.3 & 44.0 & 66.8 & \textbf{60.0} \\
\cmidrule(lr){2-14}
& \ours (Ours) & \textbf{8.2} & \textbf{21.6} & \textbf{51.0} & \textbf{20.2} & \textbf{3.08} & \textbf{11.5} & \textbf{43.9} & \textbf{8.0} & \textbf{32.8} & \textbf{49.4} & \textbf{67.8} & 59.7 \\
\bottomrule

\end{tabular}%
}
\caption{Quantitative results on generalized CZSL in \textit{open world}, all numbers are reported in percentage. S and U refer to best seen and unseen accuracy on the curve. CLIP-ZS refers to the vanilla CLIP model without fine-tuning. All CLIP-based models are run with ViT-L/14. Note that our models tested have identical weights as in Tab.~\ref{tab:closed}. \dag We run Co-CGE with similar CLIP features and report our best number of the model. Models published before CGE are omitted as their performances are inferior to current baselines.} 
\vspace{-10pt}
\label{tab:open}
\end{table*}

\textbf{Datasets.} We evaluate \ours on four popular datasets: MIT-States \cite{isola2015discovering}, C-GQA \cite{naeem2021learning}, UT-Zappos  \cite{finegrained2014,semjitter2017} and VAW-CZSL \cite{saini2022disentangling}. For splits, we follow \cite{naeem2021learning} for C-GQA, \cite{saini2022disentangling} for VAW-CZSL, and \cite{purushwalkam2019task} for MIT-States/UT-Zappos. Statistically, the numbers of images in train/val/test are 29k/10k/10k for MIT-States, 23k/3k/3k for UT-Zappos, 26k/7k/5k for C-GQA, and 72k/10k/10k for VAW-CZSL.

\textbf{Scenarios.} We perform evaluation of CZSL models on both \textit{closed} and \textit{open} world scenarios and denote them as \closedset{} and \openset{}, respectively. Regarding the \textit{closed world} setting, we follow \cite{naeem2021learning,purushwalkam2019task,atzmon2020causal} and conduct CZSL with a limited search space. We further run models in the \textit{open} world scenario, proposed by Mancini \etal \cite{mancini2021open}, to assess the scalability of CZSL models. It is worth noting that C-GQA becomes much more challenging under the \textit{open world} setting, as the size of the search space drastically increases from 2k to nearly 400k. We also notice similar space expansions on MIT-States, while the number of possible compositions does not increase much on UT-Zappos.

\textbf{Evaluation Metrics.} Our evaluation follows the generalized CZSL protocol adopted by \cite{naeem2021learning,purushwalkam2019task,atzmon2020causal, mancini2021open}. \cite{purushwalkam2019task,xian2018zero} argue that it is unreasonable to evaluate only $Y_{u}$ as significant biases enter during training and model selection. They suggest computing both seen and unseen accuracy with various bias values added to unseen categories and taking the Area Under the Curve (AUC) as the core metric. We select our models with the best AUC on \textit{val} sets and report performance on \textit{test} sets.

Furthermore, best-seen accuracy and best-unseen accuracy are calculated when other candidates are filtered out by specific bias terms. We also report best \textit{Harmonic Mean} (HM), defined as $(2*seen*unseen)/(seen+unseen)$.

\textbf{Implementation Details:} We build our model on the PyTorch \cite{paszke2019pytorch} framework. As for optimization, we use Adam optimizer with a weight decay of $5\mathrm{e}-5$. The learning rate is set to $2\mathrm{e}-5$. The batch size is set to 32 for all three datasets. The temperature $\tau_{C}, \tau_{A}, \tau_{O}$ is set to 0.01, 0.0005 and 0.0005, respectively. Most of the experiments are run on two NVIDIA A100 GPUs. We the number of vision MoA layers $M$ to 6 by default. For the downsampling function $f_{Down}$, we set the reduction factor to 4. Ablation studies on these settings can be found in Sec~\ref{sec:ablation}.

\begin{table}[t]
\centering

\resizebox{0.94\linewidth}{!}{%
\begin{tabular}{c|r|r|r|r}\toprule
\multirow{2}{*}{\shortstack{Image\\Encoder}}& Closed World & \multirow{2}{*}{\closedset{}MIT-States}&  \multirow{2}{*}{\closedset{}C-GQA} &  \multirow{2}{*}{\closedset{}UT-Zappos}\\
& Model & & &  \\ 
\midrule
\multirow{5}{*}{ViT B/32} & CLIP-ZS* \cite{radford2021learning} &  7.5 & 1.2 & 2.4 \\
& CLIP-FT \cite{radford2021learning} &  10.9 & \underline{7.6} & 21.1 \\
& Co-CGE$^\dag$ \cite{mancini2022learning} & 12.2 & 5.0  & \underline{31.2} \\
& CSP* \cite{nayak2023learning} & 12.4 & 5.7 & 24.2 \\
& DFSP \cite{lu2023decomposed} & \underline{13.2} & - & 23.3 \\
& \ours (Ours) & \bf 16.1 & \bf 10.4 & \bf 39.0 \\
\midrule
& \multicolumn{1}{c|}{\multirow{2}{*}{$\Delta$}} & +2.9 & +2.8 & +7.8 \\
& & \it (21.9\%) & \it (36.8\%) & \it (25.0\%) \\
\midrule
\multirow{7}{*}{ViT L/14} & CLIP-ZS* \cite{radford2021learning} & 11.0 & 1.4 & 5.0 \\
& CLIP-FT* \cite{radford2021learning} & 14.4 & \underline{10.5} & 4.8 \\
& CoOp* \cite{zhou2022learning} & 13.5 & 4.4 & 18.8 \\
& CLIP-Adapter* \cite{gao2021clip} & 9.5 & 3.2 & 31.5 \\
& Co-CGE$^\dag$ \cite{mancini2022learning} & 17.0 & 5.7 & \underline{36.3} \\
& CSP* \cite{nayak2023learning} & 19.4 & 6.2 & 33.0 \\
& DFSP \cite{lu2023decomposed} & \underline{20.6} & 10.5 & 36.0 \\
& \ours (Ours) &  \textbf{23.4} & 
\textbf{14.8} & 
\textbf{44.1} \\
\midrule
& \multicolumn{1}{c|}{\multirow{2}{*}{$\Delta$}} & +2.8 &  +4.3 & +7.8 \\
&  & \it (13.6\%) & \it (41.0\%)  & \it (21.5\%) \\
\bottomrule
\end{tabular}%
}
\caption{Comparison of the AUC performance on all three benchmarks among CLIP-based models. ZS and FT stand for zero-shot and fine-tuned. Best results are shown in \textbf{bold} and runner-ups are \underline{underlined}. $\Delta$ is calculated between \ours and the second-best. Numbers with * are acquired from the CSP paper \cite{nayak2023learning}. \dag We obtain these numbers by running Co-CGE on similar CLIP features.} 
\label{tab:clip}
\vspace{-8pt}
\end{table}

\subsection{Quantitative Results}
\label{sec:quan}
In this section, we present quantitative results in detail under both \textit{closed world} and \textit{open world} settings. Such results verify the effectiveness of our method, which surpasses the current SOTA on most metrics, in both scenarios.

\textbf{Closed World Results.} Performance of the \textit{closed world} scenario are reported in Tab.~\ref{tab:closed} and ~\ref{tab:vaw-czsl}. On MIT-States, results show that CAILA overcomes the label noise and achieves SOTA. More specifically, on AUC, we observe a 2.8\% improvement, from 20.6\% to 23.4\%. Furthermore, regarding HM, \ours achieves 39.9\%, outperforming all baselines. When it comes to best seen and unseen accuracy, our model improves by \ttilde 4\% and \ttilde 2\%, respectively.

Our results on C-GQA further verify the advantage of CAILA, especially when the number of unseen compositions is larger. On AUC, our model achieves a 4.3\% improvement, 40\% of the previous SOTA, from 10.5\% to 14.8\%. HM is also improved by 5.6\%. Moreover, improvements of best seen and unseen accuracy are 5.7\% and 5.6\%.

UT-Zappos has much fewer attributes and object categories, compared with its counterparts, and is thus much easier, as the gap between various methods is smaller. But it is noticeable that our model, \ours, outperforms all other baselines, with a 7.2\% improvement on the AUC metric.

Moreover, on the recently released benchmark, VAW-CZSL, \ours{} is able to achieve noticeable improvements against baseline models, particularly the newly published method, DFSP \cite{lu2023decomposed}. \ours improves the AUC by 3.1\% while boosting the harmonic mean by 3.5\%.

\textbf{Open World Results.} We further conduct experiments under the \textit{open world} setting to evaluate the robustness of \ours. Results are shown in Tab.~\ref{tab:open}. Noticeably, \textit{open world} is much harder than \textit{closed world}, as performance on all benchmarks drops drastically, while \ours achieves SOTA on most metrics in this scenario without any filtering techniques adopted in the previous papers \cite{mancini2021open, mancini2022learning, nayak2023learning}.

On MIT-States, our approach greatly beats SOTA on all metrics, particularly the AUC. Our model improves AUC from 6.8\% to 8.2\% and achieves a 21.6\% harmonic mean. Moreover, \ours achieves improves seen accuracy by 3.5\% and unseen accuracy by 0.2\%. 

The performance of \ours on C-GQA in the \textit{open world} scenario is consistent with the one in \textit{closed world}. More specifically, our model achieves 3.08\% AUC, 128\% of DFSP \cite{lu2023decomposed}. We also observe a $\sim$10\% relative improvement on harmonic mean, from 10.4\% to 11.5\%.  \ours achieves 5.6\% and 0.8\% boosts on seen and unseen.

Regarding UT-Zappos, our model also brings in performance gains. It achieves a 49.4\% harmonic mean, 4.1\% higher than Co-CGE. \ours{} also gets the best AUC of 32.8\%, at least 2\% higher against other baselines.

\textbf{Comparisons between CLIP-based Methods.} We further make head-to-head comparisons between \ours and other approaches built with CLIP in Tab.~\ref{tab:clip}, with variations on the vision encoder: ViT-B/32 \cite{dosovitskiy2021an} and ViT-L/14. Results verify \ours's effectiveness and consistency with different visual backbones. In particular, \ours achieves $>$35\% relative improvements on C-GQA against other baselines. 

\textbf{Discussion.} Given that CLIP is trained on a web-scale dataset, ensuring fair comparisons between CLIP-based \cite{lu2023decomposed,mancini2022learning,nayak2023learning} and CLIP-free methods \cite{karthik2022kg,mancini2021open,naeem2021learning} can be difficult, particularly as CLIP-based methods significantly outperform CLIP-free ones. We follow the setting in existing CLIP-based methods \cite{lu2023decomposed,mancini2022learning,nayak2023learning,radford2021learning,zhou2022learning} with a focus on enhancing CLIP-based CZSL. Comparisons between \ours and fine-tuned CLIP models show that a partially tuned model can beat its fully fine-tuned counterpart by a large margin, justifying that \ours{} better suppresses training biases while remaining sharp on knowledge transfer for CZSL, thus is a better way to exploit CLIP knowledge. 

\begin{figure*}[t]
     \centering
     \hfill
     \begin{subfigure}[b]{0.33\linewidth}
         \centering
         \includegraphics[width=\textwidth]{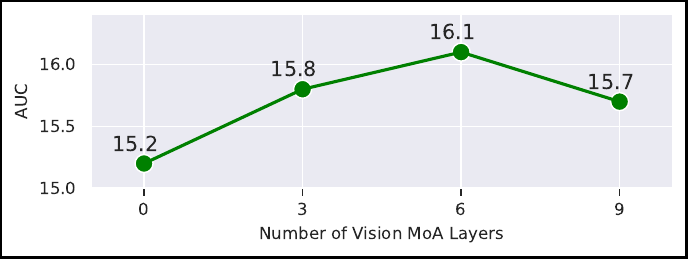}
         \caption{Number of Vision MoA Layers $M$}
         \label{tab:abl-moa}
     \end{subfigure}
     \begin{subfigure}[b]{0.33\linewidth}
         \centering
         \includegraphics[width=\textwidth]{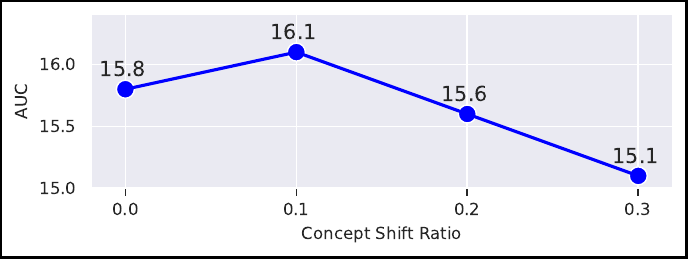}
         \caption{Concept Shift Ratio}
         \label{tab:abl-shift}
     \end{subfigure}
     \begin{subfigure}[b]{0.33\linewidth}
         \centering
         \includegraphics[width=\textwidth]{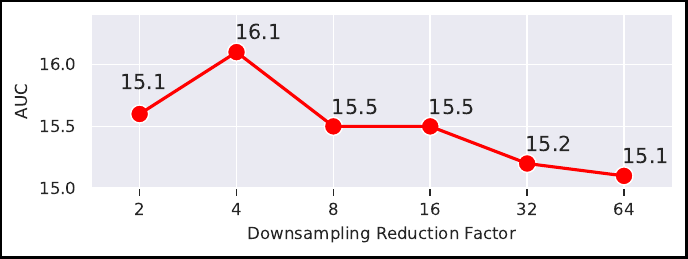}
         \caption{Reduction Factor}
         \label{tab:abl-red}
     \end{subfigure}
        \caption{Ablation studies: (a) The number of vision MoA layer $M$; (b) The ratio of concept shift; (c) The reduction factor of $f_{Down}$. }
        \label{fig:linechart}
        \vspace{-10pt}
\end{figure*}

\subsection{Ablation Studies}

We conduct the ablation study with CLIP ViT-B/32 and MIT-States in \textit{closed world}.

\textbf{Adapter and MoA. } We evaluate different adapter/MoA settings on MIT-States and report results in Tab.~\ref{tab:adap-moa}. We observe that compared with CSP \cite{nayak2023learning}, adding adapters to either side of encoders can effectively improve the performance while attaching adapters to both sides shows further improvements. Experiments in the last three rows verify that our Mixture-of-Adapters mechanism further improves the performance when it is applied on both sides.

\textbf{Vision Mixture Strategies. }We compare different ways of mixing $\vec{z}$ and $\vec{h’}$ inside the vision MoA layer as described in Eqn.~\ref{eqn:mix-z},\ref{eqn:mix-h}. Tab.~\ref{tab:abl-mix} shows the results of mixing only one of the feature vectors or none at all. The last row corresponds to averaging $\mathcal{F}_A(x), \mathcal{F}_O(x), \mathcal{F}_C(x)$ without intra-layer mixture, which is similar to the language side MoA. Experiment results demonstrate that mixing both $\vec{z}$ and $\vec{h'}$ as proposed in Sec.~\ref{sec:moa} yields optimal performance while applying a similar strategy as the language side hurts.

\begin{table}[t]
\centering\footnotesize
\begin{tabular}{cc|cc|rrrr} \toprule
 \multicolumn{2}{c|}{Adapter} & \multicolumn{2}{c|}{MoA} & \multicolumn{4}{c}{\closedset{} MIT-States} \\
 V & L & V & L & AUC ($\uparrow$) & HM ($\uparrow$)  & S ($\uparrow$)   & U ($\uparrow$)   \\ 
 \midrule
 \multicolumn{4}{c|}{CSP \cite{nayak2023learning}} & 12.4 & 28.6 & 36.4 & 42.5 \\
\midrule
\cmark &  & & & 14.0 & 30.1 & 41.4 & 42.0 \\
& \cmark & & & 13.9 & 30.5 & 40.3 & 42.8 \\
\cmark & \cmark & & & 14.4 & 30.7 & 42.2 & 43.2 \\
\cmark & \cmark & \cmark &  & 15.4	 & 31.4 & \bf 43.4 & 44.5\\
\cmark & \cmark & & \cmark & 15.2 & 31.7 & 41.6 & 44.8 \\

\midrule
 \cmark & \cmark & \cmark & \cmark &\bf 16.1 & \bf 32.9 & 43.3 & \bf 45.6\\
\bottomrule
\end{tabular}

\caption{Ablation on adapters and MoA modules. V and L refer to Vision and Language, respectively.}
\label{tab:adap-moa}
\end{table}

\begin{table}[t]
\centering\footnotesize
\resizebox{0.92\linewidth}{!}{
\begin{tabular}{r|cc|rrrr} \toprule
Closed World & \multicolumn{2}{c|}{Mixture} & \multicolumn{4}{c}{\closedset{} MIT-States} \\
 Model & $\vec{z}$ & $\vec{h'}$ & AUC ($\uparrow$) & HM ($\uparrow$)  & S ($\uparrow$)   & U ($\uparrow$)   \\ 
\midrule
\multirow{4}{*}{CAILA (Ours)} & \cmark & \cmark &\bf 16.1 & \bf 32.9 & \bf 43.3 & \bf 45.6\\
\cmidrule{2-7}
& \cmark &  & 15.8 & 32.2 & 43.3 & 45.2 \\
& & \cmark & 15.5 & 31.7 & 43.0 & 45.1\\
& & & 15.5 & 32.0 & 42.7 & 44.8 \\
\bottomrule
\end{tabular}
}
\caption{Ablation on vision MoA strategies.}
\label{tab:abl-mix}
\vspace{-10pt}
\end{table}  

\textbf{Vision Mixture Functions.} We evaluate various mixture functions of vision MoA besides the default mean function, including summation (Sum.), element-wise multiplication (Mul.), and concatenation (Concat.). We add one linear layer after ``Concat'' to align the feature dimension with upcoming operations. Results in Tab.~\ref{tab:abl-mix-fn} show that the ``Mean'' operation performs the best. We also notice that the variation with "Sum." performs worse, possibly because summation greatly changes the magnitude of the feature vector.

\textbf{Learnable Prompts.} We perform experiments to study the effect of learnable prompts in our framework. Results reported in Tab.~\ref{tab:abl-prompt} show that our model remains competitive with prompt embeddings fixed. Such behavior justifies that performance gains of \ours{} come from designs that have been discussed in the Approach section.

\begin{table}[t]
\centering\footnotesize
\resizebox{0.92\linewidth}{!}{
\begin{tabular}{r|r|rrrr} \toprule
Closed World & \multicolumn{1}{c|}{\multirow{2}{*}{Mix. Fn.}} & \multicolumn{4}{c}{\closedset{} MIT-States} \\
 Model & & AUC ($\uparrow$) & HM ($\uparrow$)  & S ($\uparrow$)   & U ($\uparrow$)   \\ 
\midrule
\multirow{4}{*}{CAILA (Ours)} & Mean & \bf 16.1 & \bf 32.9 & \bf 43.3 & \bf 45.6 \\
\cmidrule{2-6}
& Sum. & 14.6 & 30.7 & 42.8 & 42.1 \\
& Mul. & 15.8 & 32.2 & 43.3 & 45.0 \\
& Concat. & 15.2 & 31.9 & 41.8 & 44.8\\
\bottomrule
\end{tabular}
}
\vspace{-3pt}
\caption{Ablation on vision MoA mixture functions.}
\label{tab:abl-mix-fn}
\vspace{-7pt}
\end{table}

\begin{table}[t]
\centering\footnotesize
\begin{tabular}{r|rrrr} \toprule
Closed World & \multicolumn{4}{c}{\closedset{} MIT-States} \\
 Model & AUC ($\uparrow$) & HM ($\uparrow$)  & S ($\uparrow$)   & U ($\uparrow$)   \\ 
 \cmidrule(r){1-1} \cmidrule(l){2-5}
CAILA(Ours) & \bf 16.1 & \bf 32.9 & 43.3 & \bf 45.6\\
\cmidrule(r){1-1} \cmidrule(l){2-5}
w/o Learnable Prompts & 15.8 & 32.1 & \bf 43.5 & 44.6\\
DFSP \cite{lu2023decomposed} & 13.2 & 29.4 & 36.7 & 43.4 \\
CSP \cite{nayak2023learning} & 12.4 & 28.6 & 36.4 & 42.5 \\
\bottomrule
\end{tabular}
\caption{Ablation study on learnable prompts.}
\label{tab:abl-prompt}
\vspace{-15pt}
\end{table}

\label{sec:ablation}

\textbf{\ours Setups.} We explore different aspects of our setup and show the results in Fig.~\ref{fig:linechart}. Fig.~\ref{fig:linechart}(a) demonstrates that \ours performs better with MoA layers and achieves the best performance with 6 MoA layers on the vision side, which is also better than the single-stage MoA when $M$=0; Fig.~\ref{fig:linechart}(b) indicates that replacing 10\% of a batch with post-shift features can increase the AUC while adding more shift reduces it; In Fig.~\ref{fig:linechart}(c), we find that the optimal reduction factor for the latent feature $\vec{z}$ is 4, while using higher reduction factors does not affect the performance significantly and can be considered for efficiency reasons.

\section{Conclusion}
In this paper, we explore the problem of how to leverage large-scale Vision-Language Pre-trained (VLP) models, particularly CLIP, more effectively for compositional zero-shot learning. Unlike previous methods which treat CLIP as a black box, we propose to slightly modify the architecture and attach Concept-Aware Intra-Layer Adapters (CAILA) to each layer of the CLIP encoder to enhance the knowledge transfer from CLIP to CZSL. Moreover, we design the mixture-of-adapters mechanism to further improve the generalizability of the model. Quantitative evaluations demonstrate that \ours achieves significant improvements on all three common benchmarks. Due the lack of unfeasible pair filter, CAILA's performance drops from closed world to open world, when the number of possible pairs greatly increases, though. We also provide comprehensive discussions on deciding the optimal setup.

\section*{Acknowledgment}
This research was supported, in part, by the Office of Naval Research under grant \#N00014-21-1-2802.

{\small
\bibliographystyle{ieee_fullname}
\bibliography{egbib}
}

\end{document}